\documentclass[10pt,twocolumn,letterpaper]{article}
\usepackage[pagenumbers]{cvpr} % To force page numbers, e.g. for an arXiv version
\definecolor{cvprblue}{rgb}{0.21,0.49,0.74}
\usepackage[pagebackref,breaklinks,colorlinks,allcolors=cvprblue]{hyperref}
\usepackage[capitalize]{cleveref}

\def\name{EventGPT}
\newcommand{\bfsection}[1]{\vspace*{0.0mm}\noindent\textbf{#1.}}

\def\paperID{9914} % *** Enter the Paper ID here
\def\confName{CVPR}
\def\confYear{2025}

\title{EventGPT: Event Stream Understanding with Multimodal \\ Large Language Models}

\author{Shaoyu Liu$^{1}$, Jianing Li$^{2}$, Guanghui Zhao$^{1}$, Yunjian Zhang$^{3}$, Xin Meng$^{4}$\\ Fei Richard Yu$^{5}$, Xiangyang Ji$^{2}$, Ming Li$^{5}$ \\
\\
$^{1}$Xidian University ~ $^{2}$Tsinghua University ~ $^{3}$UCAS ~ $^{4}$Peking University \\
$^{5}$Guangdong Laboratory of Artificial Intelligence and Digital Economy(SZ)
}

\begin{document}
\maketitle
\begin{abstract}
% Event cameras capture visual information as asynchronous event streams, providing advantages such as high temporal resolution and low latency. However, their sparse and dynamic nature poses significant challenges for traditional vision models, which are designed for dense, frame-based data. Existing multimodal large language models excel at processing natural images but struggle with the unique characteristics of event data. In this paper, we introduce EventChat, the first multimodal large language model based on event-camera data, aimed at bridging the gap between event streams and language understanding. To address the inherent modal differences and the spatiotemporal complexities of event data, we propose a feature adaptor to reconcile cross-modal disparities and incorporate a spatiotemporal processing module during feature extraction, enhancing the model’s capacity to handle temporal sequences and spatial changes. Additionally, we construct a large-scale Event-Text dataset with 1000k pre-training pairs and 120k instruction-following samples, facilitating effective learning and generalization in the event modality. Experimental results demonstrate that EventChat effectively understands and reasons across a wide range of event stream data, serving as a prototype for future chat-centric event stream understanding research.
Event cameras record visual information as asynchronous pixel change streams, excelling at scene perception under unsatisfactory lighting or high-dynamic conditions. Existing multimodal large language models (MLLMs) concentrate on natural RGB images, failing in scenarios where event data fits better.
% However, their sparse and dynamic nature poses significant challenges for traditional vision models, which are designed for dense, frame-based data.  
% In this paper, we introduce the first MLLM \name{} for event stream understanding, to the best of our knowledge, bridging the gap between event streams and natural language.
In this paper, we introduce \name{}, the first MLLM for event stream understanding, to the best of our knowledge, marking a pioneering attempt to integrate large language models (LLMs) with event stream comprehension.
% Generally, we leverage LLMs as an anchor to align the event modality with language as the pre-trained LLMs possess massive knowledge about the world and provide a powerful foundation for cross-modal alignment. 
To mitigate the huge domain gaps, we develop a three-stage optimization paradigm to gradually equip a pre-trained LLM with the capability of understanding event-based scenes.
Our \name{} comprises an event encoder, followed by a spatio-temporal aggregator, a linear projector, an event-language adapter, and an LLM.
% Stage One employs the LLaVA-pre-train dataset to establish critical visual-language associations, laying the groundwork for robust multimodal alignment. 
Firstly, RGB image-text pairs generated by GPT are leveraged to warm up the linear projector, referring to LLaVA, as the gap between natural image and language modalities is relatively smaller. 
% Stage Two leverages our custom-built N-ImageNet-Chat dataset, finely tuning the model’s feature mapping to enable effective cross-modal transfer of visual knowledge into the event domain. 
Secondly, we construct a synthetic yet large dataset, N-ImageNet-Chat, consisting of event frames and corresponding texts to enable the use of the spatio-temporal aggregator and to train the event-language adapter, thereby aligning event features more closely with the language space. Finally, we gather an instruction dataset, Event-Chat, which contains extensive real-world data to fine-tune the entire model, further enhancing its generalization ability.
%Finally, we gather a high-quality real-event dataset, Event-Chat, to perform instructional fine-tuning of the entire model, further enhancing its generalization ability.
% This progressive approach endows the model with enhanced temporal awareness and a sophisticated capability to manage complex, multimodal interactions across visual and event domains.
% 接下来说验证实验
% We construct a comprehensive evaluation benchmark, and extensive experiments demonstrate that \name{} outperforms previous state-of-the-art MLLMs in generation quality, descriptive accuracy, and reasoning capability.
We construct a comprehensive benchmark, and experiments show that \name{} surpasses previous state-of-the-art MLLMs in generation quality, descriptive accuracy, and reasoning capability. Code: \href{https://xdusyl.github.io/eventgpt.github.io/}{\textbf{EventGPT}}
% Our code and datasets will be available at \textbf{\url{https://github.com/XduSyL/EventGPT}}
% Code: \href{https://github.com/XduSyL/EventGPT}{\textbf{EventGPT}}

% we choose to align the event features with language embedding space due to the and the powerful capacity of LLMs.

% To address the inherent modal differences and the spatiotemporal complexities of event data, we propose a feature adaptor to reconcile cross-modal disparities and incorporate a spatiotemporal processing module during feature extraction, enhancing the model’s capacity to handle temporal sequences and spatial changes. Additionally, we construct a large-scale Event-Text dataset with 1000k pre-training pairs and 120k instruction-following samples, facilitating effective learning and generalization in the event modality. Experimental results demonstrate that EventGPT effectively understands and reasons across a wide range of event stream data, serving as a prototype for future chat-centric event stream understanding research.
\end{abstract}    
\begin{figure}[t]
  \centering
   \includegraphics[width=\linewidth]{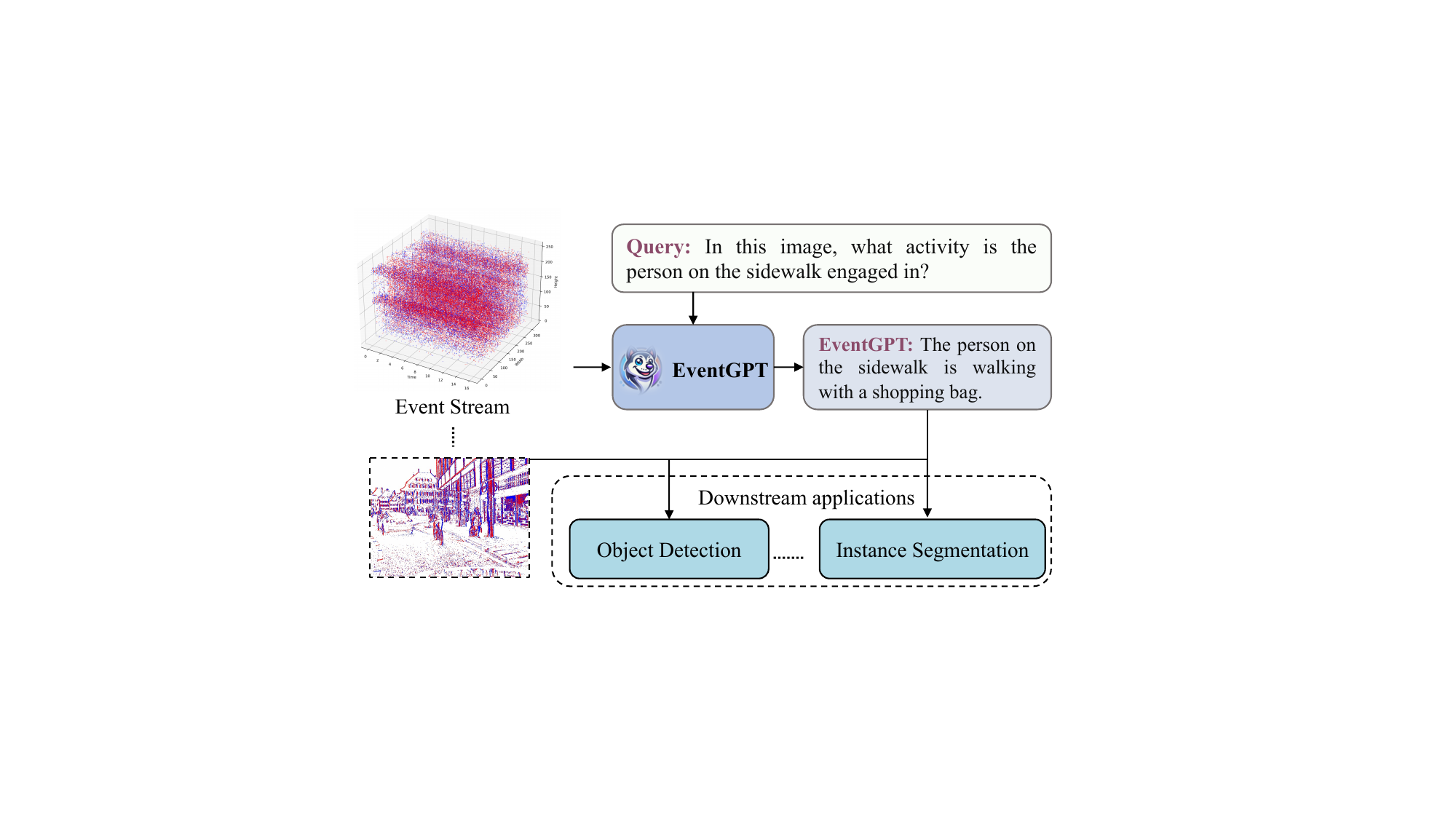}
    \caption{Our \name{} is the \textit{first} multimodal large language model tailored for event stream understanding, including scene summarization, reasoning, and question answering, with a great potential for downstream tasks.}
   %\caption{EventGPT: Event-driven perception and understanding in event streams.}
   \label{fig:motivation}
\end{figure}

\section{Introduction}

\begin{figure*}[t]
  \centering
   \includegraphics[width=\linewidth]{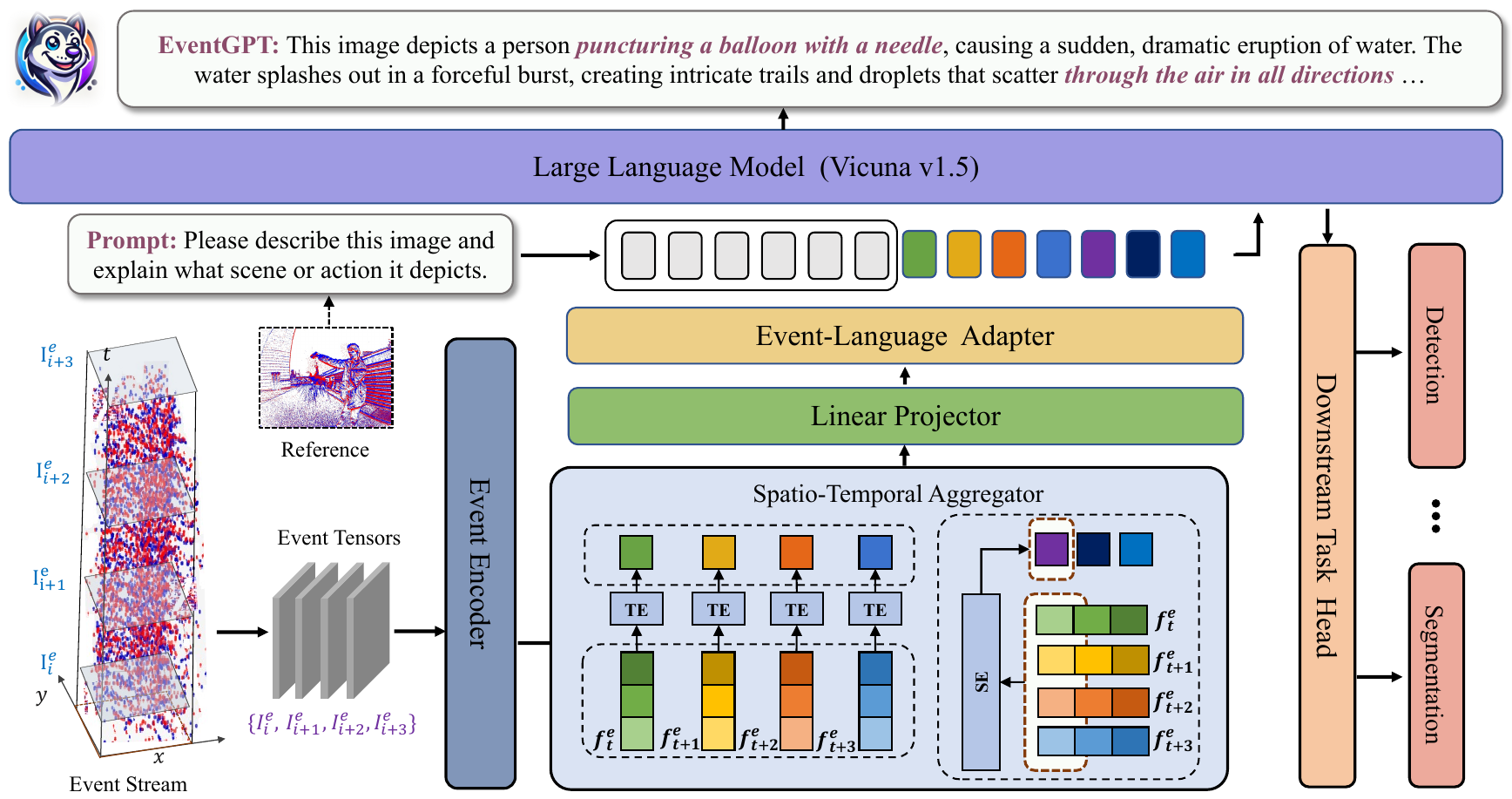}
   \caption{
   % Overview of our framework. The event encoder processes raw event tensors into high-dimensional features, which are aggregated using a spatio-temporal aggregator. These representations are projected and aligned with the large language model, achieving a nuanced understanding of event streams, with potential applications in downstream tasks.
   Overview of our framework. The event encoder transforms raw event tensors into high-dimensional features, which are then aggregated by a spatio-temporal module. These representations are projected and aligned with the large language model, enabling nuanced understanding of event streams and supporting various downstream applications.
   }
   \label{fig:overview}
\end{figure*}

Event cameras, with predominant advantages in capturing scene changes and temporal resolution, have great application potential in scenarios under imperfect lighting conditions or including high dynamic objects \cite{gallego2020event, zheng2302deep, huang2023event}. 
%
%These unique characteristics have driven advancements in various downstream tasks, such as object detection, tracking, and 3D reconstruction, propelling progress in cutting-edge computer vision applications.
The relevant tasks on event data, \eg object detection \cite{gehrig2023recurrent, li2023sodformer, gehrig2024low, wu2024leod}, tracking \cite{messikommer2023data, zhang2022spiking, shen2024blinktrack, gehrig2020eklt}, and 3D reconstruction \cite{yu2024eventps, muglikar2021esl, kim2016real, rebecq2018emvs}, have been well-investigated. 

% combines visual features with large-scale language models to create a system capable of understanding and generating multimodal content, significantly improving performance on visual question answering and vision-language generation tasks. 
Recent multimodal large language models (MLLMs) have shown remarkable breakthroughs in conventional vision and language problems, demonstrating extraordinary capabilities of scene understanding and question answering (QA) \cite{liu2024visual, zhu2023minigpt, xue2024xgen, liu2024improved, shen2024aligning}. Typically, a pioneering research Large Language and Vision Assistant (LLaVA) combines a vision encoder with an LLM, trained on GPT-4-generated multimodal data, to create a robust system that excels in visual understanding, instruction-following, and complex question answering \cite{liu2024visual}.  
% presenting a promising opportunity to leverage the distinctive spatiotemporal information captured by event cameras.
These works mainly pay attention to natural RGB scenes and the resulting models degrade significantly when encountering the aforementioned scenarios. However, the relevant research on event data streams remains underexplored.

% Recent efforts have begun to align and integrate event-based data with textual information, as exemplified by pioneering approaches like EventCLIP and EventBind. 
Some works have initialized the alignment between event data and textual information. For example, EventCLIP \cite{wu2023eventclip} adapts the pre-trained CLIP model \cite{radford2021learning, cherti2023reproducible} for event-based object recognition, enabling zero-shot and few-shot classification. Similarly, EventBind \cite{zhou2023clip} extends CLIP with an event encoder to align images, texts, and events in a unified representation space, realizing few-shot recognition and event retrieval. 
% While these models represent significant strides in multimodal learning, they are limited in their capacity to fully leverage the rich spatiotemporal features inherent in event-based data and often fall short in providing robust contextual understanding in complex scenarios. 
Nevertheless, both EventCLIP and EventBind lack the ability to leverage the embedded world knowledge in LLMs, limiting their capacity for nuanced scene understanding and dialog generation in complex scenarios.
%This gap underscores the necessity for a model like EventGPT, specifically designed to address these limitations by harnessing the advanced multimodal integration capabilities of large language models.
This gap highlights the necessity of a model, akin to LLaVA for RGB images, specifically designed to overcome these limitations by harnessing the power of LLMs for comprehensive event stream understanding.

% EventGPT aims to achieve a more comprehensive fusion of event-based information and textual data, thereby enhancing contextual understanding across diverse and complex scenarios.
In this work, we pioneer the development of an MLLM for event stream understanding, termed \name{} (see Fig~\ref{fig:motivation}), designed for the community to support scene summarization, reasoning, and question answering. 
% However, several key challenges remain.
To this end, several key challenges must be addressed.
% due to modality disparities, spatiotemporal complexities, and limited data resources. 
% First, the fundamental differences between event data, which captures brightness changes through sparse, event-based information that lacks complete visual context, and conventional image modalities complicate the direct adaptation of multi-modal techniques typically designed for natural images. 
% First of all, event streams inherently lack global visual context, as they are designed to capture asynchronous and sparse brightness changes within the field of view, making it significantly challenging to align with the high-level semantics of natural language. 
%三个挑战
%1. 单独的事件缺乏全局语义，需要与NL集合
%2. 直接用现有的MLLMs不行，需要设计专门的MLLMs
%3. 数据集
First of all, aligning event features with the high-level semantics of natural language is significantly challenging, as event streams are designed to capture asynchronous and sparse brightness changes within the field of view, inherently lacking the global scene context necessary for language-based understanding.
Secondly, the fundamental difference in how event and RGB cameras capture scene information complicates the straightforward adaptation of existing multimodal techniques, which are typically designed for natural images.
% Second, event data contains rich spatiotemporal information, necessitating models capable of effectively processing temporal sequences. However, current multimodal large language models, such as LLaVA and MiniGPT-4, are optimized primarily for static image understanding. While temporal models like VideoGPT and Video-LLaMA are suitable for sequence processing, they introduce significant computational overhead, limiting efficiency for continuous event stream processing. 
% Finally, the scarcity of large-scale, high-quality Event-Text multimodal datasets restricts the models’ capacity to learn and generalize in the event modality, as most available datasets are unimodal and lack the annotations necessary for effective multimodal learning.
Finally, the scarcity of large-scale, high-quality event-text datasets hinders the training of a scalable MLLM for comprehensive event scene understanding.

To address these challenges, we propose a three-stage training paradigm that leverages an LLM as an anchor, given its rich world knowledge and a strong foundation for scene understanding, to progressively align the event modality with language. In general, our \name{} is mainly composed of an event encoder, a spatio-temporal aggregator, a linear projector, an event-language adapter, and an LLM.
Compared to event data, natural images are much closer to language in the embedding space. Therefore, in the first stage, we use the image-text dataset from LLaVA to warm up the linear projector, establishing preliminary scene-language associations. 
%Stage Two bridges the cross-modal domain gap with an Event Feature Adaptor, transferring visual knowledge into the event domain. Given the scarcity of Event-Text data, we construct a large-scale Event-Text dataset, N-ImageNet-Chat, to support this stage. 
% In the second stage, we aim to train the spatio-temporal aggregator and event-language adapter to adapt the spatio-temporal information from event frames to language space. To achieve it, we construct a synthetic large-scale event-text dataset, N-ImageNet-Chat, to support the optimization. 
% In the second stage, we train the spatio-temporal aggregator and event-language adapter to map spatio-temporal information from event frames into language space. To accomplish this, we construct a synthetic large-scale event-text dataset, N-ImageNet-Chat, to support optimization.
In the second stage, we focus on training the spatio-temporal aggregator and event-language adapter to map spatio-temporal information from event frames into language space. For this purpose, we construct a large-scale synthetic event-text dataset, N-ImageNet-Chat, to facilitate optimization.
%In Stage Three, we perform full fine-tuning across all parameters, allowing the model to interpret event stream data with refined contextual awareness. For this stage, we created an instruction-supervised extension of N-ImageNet-Chat and developed the Event-Chat dataset,
% along with introducing a spatiotemporal module that aggregates features by separately applying mean pooling across both spatial and temporal dimensions, enabling the model to capture complex spatiotemporal dynamics.
Finally, we conduct instructional fine-tuning of all model parameters on a self-collected, high-quality dataset, Event-Chat, which includes diverse scenes with unsatisfactory lighting conditions or high-speed object motions, enhancing the capability of our \name{} to interpret complex real-world scenes.

We introduce a comprehensive evaluation benchmark inspired by Qwen2-72B-Instruct to assess the generation quality, descriptive accuracy, and reasoning capability of \name{}. To facilitate a rigorous comparison, we select several state-of-the-art MLLMs excelling in natural scene understanding, including LLaVA \cite{liu2024visual}, Qwen2-VL \cite{wang2024qwen2}, and Deepseek-VL \cite{lu2024deepseek}, as baseline models. Evaluation is conducted across three metrics, \ie Detail Captioning (DC), Complex Reasoning (CR), and Visual Question Answering (VQA), in line with the existing literature. Experimental results highlight significant improvements achieved by \name{}. Furthermore, our model demonstrates strong generalization ability on open-set downstream tasks, such as object detection and semantic segmentation.

Our main contributions in this work can be summarized as follows:
\begin{itemize}[noitemsep,topsep=0pt,leftmargin=15pt]
    % \item We introduce \textbf{\name{}}, the first multimodal large language model for event scene understanding, representing a pioneering endeavor in creating a system focused on understanding event streams.
    \item We present \name{}, the first multimodal large language model tailored for event stream understanding, demonstrating capabilities in scene summarization, reasoning, and question answering. \name{} establishes a solid baseline, paving the way for future research. %in the community.
    % \item We propose a phased training strategy to mitigate cross-modal discrepancies and incorporate a spatiotemporal processing module during feature extraction, enhancing the model’s ability to handle complex spatiotemporal information.
    \item We propose a systematic learning paradigm incorporating specialized training strategies and network modules to progressively bridge the substantial domain gap between event data and language, providing a valuable reference for future cross-modal learning efforts.
    % \item We construct a large-scale \textbf{Event-Text dataset} containing both real-world and synthetic scenes, including 1 million samples for pre-training and 120,000 samples for instruction tuning, to facilitate effective multimodal learning specifically in the event-based domain.
    \item We construct two large-scale event-text datasets, N-ImageNet-Chat and Event-Chat, containing 1 million synthetic samples for pre-training and 120,000 instruction samples for fine-tuning. These datasets will be publicly released to drive progress in multimodal learning within the event domain.
\end{itemize}

%-------------------------------------------------------------------------
\section{Related Works}
\label{sec:formatting}

\noindent\textbf{Multimodal Large Language Models.} The significant advancements in Large Language Models (LLMs) for natural language understanding have driven their integration into multimodal research domains, with a particular focus on enhancing vision-language alignment to expand their applicability \cite{zhang2024mm, yin2023survey, caffagni2024r, wang2024cloud, lin2024vila, zhang2024omg}.~\cite{singh2022flava, chen2024dress, shen2024aligning, chen2023first, pandey2022cross}. Early contributions, such as Flamingo~\cite{alayrac2022flamingo} and BLIP~\cite{li2022blip, li2023blip}, integrated visual and language modalities to enable zero-shot generalization across a broad range of visual-language tasks. Further advancements emerged with chat-oriented multimodal large language models(MLLMs) like MiniGPT-4~\cite{zhu2023minigpt} and LLaVA~\cite{liu2024visual}, which improved the interactive capabilities and adaptability of LLMs in multimodal applications, setting new benchmarks in performance. As MLLMs continue to evolve, the exploration of novel sensor modalities holds potential for expanding their functional scope. Despite the distinct advantages of event cameras, such as high temporal resolution and robustness in challenging lighting conditions, the integration of event-based data with LLMs remains underexplored, particularly for applications in high-dynamic and low-light environments.

\noindent\textbf{Event-Driven Multimodal Models}. The integration of event-based data into vision-language models has attracted significant interest, demonstrating potential across various downstream tasks and highlighting the advantages of event-language synergy.~\cite{wu2023eventclip, zhou2024exact, zhou2023clip}. Notably, frameworks such as ExACT have leveraged language-guided frameworks for event-based action recognition, effectively aligning action sequences with contextual textual cues to achieve strong performance~\cite{zhou2024exact}. Other approaches, such as EventCLIP~\cite{wu2023eventclip} and EventBind~\cite{zhou2023clip}, have also demonstrated strong zero-shot performance, underscoring the effectiveness of event-based embeddings when aligned with textual information. Despite these advances, developing chat-centric large language models (LLMs) capable of incorporating event-based modalities presents significant research challenges. Addressing this gap could enhance models’ contextual comprehension, enabling more effective processing of complex interactive tasks.

%------------------------------------------------------------------------
\section{Event-Text Data Generation}
\begin{figure}[t]
  \centering
   \includegraphics[width=\linewidth]{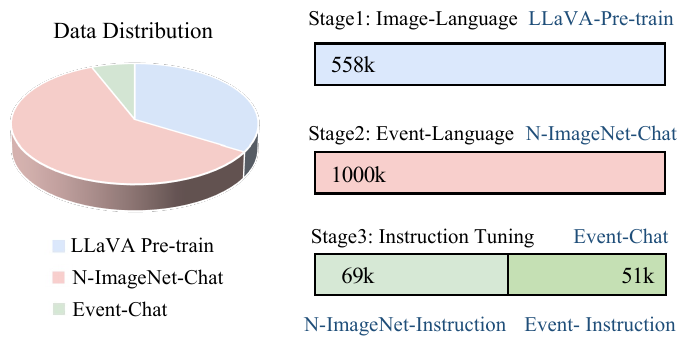}
   \caption{Data distribution across three progressive training stages in our EventGPT containing image-language alignment, event-language alignment, and instruction tuning.}
   \label{Data_Distribution}
\end{figure}
Event-based datasets have been extensively studied in computer vision tasks. However, large-scale Event-Text datasets suitable for training MLLMs) are not yet widely available. Previous studies have successfully employed large models, such as GPT, to synthesize data for multimodal applications, yielding significant advancements in model capabilities~\cite{wu2023next, jiao2024img, zhan2024anygpt, zhu2023vl}. To address this gap, we construct an Event-Text dataset using a human-assisted, semi-automated pipeline that incorporates both synthetic and real-world event data. The distribution of the dataset is shown in Fig.~\ref{Data_Distribution}.

% The computer vision community has developed a variety of event-based datasets, yet there remains a notable gap in large-scale Event-Text datasets, especially high-quality corpora tailored for training multimodal large language models (MLLMs). Previous research has successfully employed large models, such as GPT, to synthesize data for multimodal applications, yielding significant advancements in model capabilities \cite{wu2023next, jiao2024img, zhan2024anygpt, zhu2023vl}.

% In this work, we address this gap by constructing a comprehensive Event-Text dataset through a human-assisted, semi-automated pipeline. First, we sourced open-access event-image paired datasets, such as N-ImageNet\cite{kim2021n} and DSEC\cite{gehrig2021dsec}. Then, leveraging multimodal LLMs, we generated a diverse set of textual instructions for each image, including captions, visual question answering (VQA), and complex reasoning tasks. To enhance data quality in challenging scenarios, such as low-light and high-speed events, we integrated human-assisted prompts to better guide model outputs. This process resulted in a pre-training dataset of over one million instances and an instruction-supervised dataset containing over 120,000 entries, encompassing both synthetic and real-world data.

\begin{figure}[t]
  \centering
   \includegraphics[width=\linewidth]{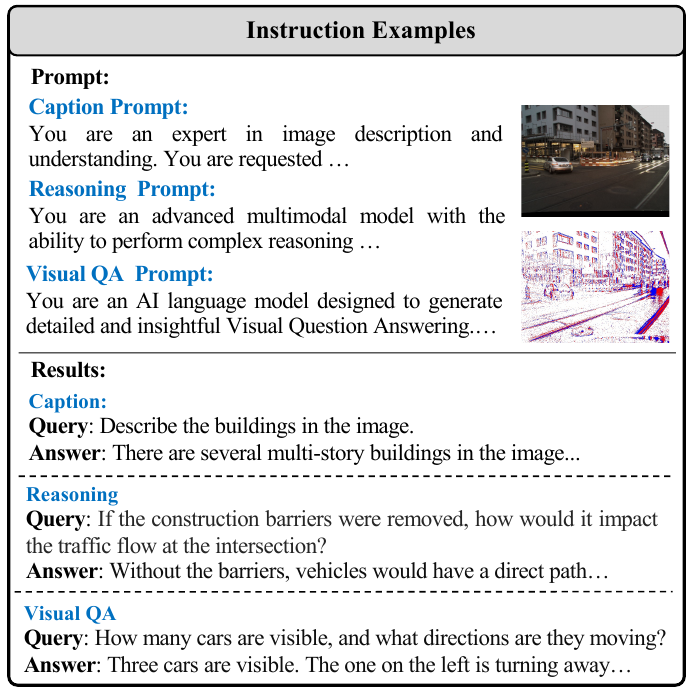}
   \caption{Instruction examples from the Event-Chat dataset for multimodal task prompts and responses in image captioning, visual reasoning, and visual question answering.}
   \label{Instruction_examples}
\end{figure}

\subsection{Synthetic Dataset}

ImageNet serves as a foundational benchmark for visual recognition tasks. N-ImageNet~\cite{kim2021n} extends this foundation by providing a synthetic event-based dataset generated through event simulators, aligned with ImageNet images. Using this alignment, we generate textual instructions for the images to enable cross-domain learning between visual and event-based data.

%Our approach produces two distinct datasets:

\noindent\textbf{N-ImageNet-Chat}. We present a large-scale dataset comprising over 1,000,000 instances with concise instructions, primarily designed to align event data with natural language. This dataset provides a foundation for the model’s understanding and supports robust representation learning.

\noindent\textbf{N-ImageNet-Instruction}. We establish an event-based dataset of over 69,000 entries, designed to improve the model’s responsiveness to complex, task-specific instructions. The dataset includes captions, visual question answering (VQA), and complex reasoning tasks aimed at fine-tuning the model’s multimodal capabilities.

% These datasets collectively form the simulated N-ImageNet-Chat, integrating foundational and task-specific knowledge for improved multimodal event-based applications.

\subsection{Real-World Dataset}
In addition to synthetic datasets, we establish a real-world dataset to improve model generalization for practical applications. A human-assisted and semi-automated annotation process ensures high-quality labels for complex scenes.

\noindent\textbf{Event Data for Event-Chat}. We leverage real-world event datasets, such as DSEC~\cite{gehrig2021dsec} and E2VID~\cite{Rebecq19cvpr}, which are primarily designed for autonomous driving and provide synchronized image-event data under challenging conditions, including low-light environments and high-speed motion (e.g., entering/exiting tunnels, night scenes).

\noindent\textbf{Instructions for Event-Chat}. This dataset is designed to enhance the model’s generalization capabilities, consisting of 59,000 high-quality annotated entries. Three distinct prompting strategies (i.e., captioning, VQA, and complex reasoning) form the basis of the Event-Chat dataset, providing diverse and comprehensive instructions. As shown in Fig.~\ref{Instruction_examples}, examples from each prompting strategy illustrate the dataset’s richness in contextual understanding. Combined with N-ImageNet-instruction, these annotated entries form the final Event-Chat dataset.

In summary, our newly built event-text dataset contains over 1,000,000 entries from both synthetic and real-world data, offering a robust foundation for advancing multimodal understanding in both event-based and textual domains. These fundamental and challenging datasets are intended to foster progress in the development of MLLMs within the neuromorphic community.

\definecolor{verylightgray}{gray}{0.88}
\section{EventGPT}
\subsection{Overview}
\label{sec:pipeline}

EventGPT aims to serve as an event-based MLLM capable of understanding and generating responses based on spatio-temporal event data. As illustrated in Fig.~\ref{fig:overview}, EventGPT formulates event data processing as an event-driven approach to question answering and descriptive generation. By leveraging the high temporal resolution and extended dynamic range inherent in event stream data, EventGPT enhances understanding in scenarios traditionally challenging for standard vision models, such as low-light or high-speed motion conditions. At the core of EventGPT lies an LLM that synthesizes complex spatio-temporal information from event streams, utilizing world knowledge to generate coherent, human-readable outputs.

% \noindent\textbf{Problem Definition}. 
% In general, an event bin can be mathematically described as $S$ $=$ $ \{(x_j, y_j, t_j, p_j)\}_{j=1}^N $ within a time window \( T \), where each event consists of spatial coordinates \( (x_j, y_j) \), timestamp \( t_j \), and polarity \( p_j \). We begin by extracting high-dimensional event features \( V \) from \( S \) using an event encoder \( f_{\text{vis}} \), initialized with OpenCLIP ViT-L/14-336px to leverage its robust feature extraction capabilities. These features are then processed through a spatio-temporal encoder \( f_{\text{ST}} \) to capture dynamic interactions, resulting in a comprehensive representation \( E \). Then, a two-stage projector is employed to achieve cross-modal alignment. The first stage aligns the visual features with a textual representation, and the second stage adapts this representation specifically to the event modality, resulting in an event-aligned feature \( E \). Given a user query \( Q \), the large language model \( f_{\text{llm}} \) performs reasoning over \( E \) to generate a response \( A \) as:
% \begin{equation}
% \resizebox{0.9\hsize}{!}{$
% \begin{aligned}
%     V &= f_{\text{enc}}(S), \:
%     E = f_{\text{proj}}(f_{\text{align}}(V)), \:
%     A = f_{\text{llm}}(E \mid Q).
% \end{aligned}
% $}
% \end{equation}
In general, an event bin can be mathematically described within a time window \( T \) as \( S = \{(x_j, y_j, t_j, p_j)\}_{j=1}^N \), where each event consists of spatial coordinates \( (x_j, y_j) \), timestamp \( t_j \), and polarity \( p_j \). Initially, high-dimensional event features \( V \) are extracted from \( S \) using an event encoder \( f_{\text{enc}} \), initialized with OpenCLIP ViT-L/14-336px to leverage its robust feature extraction capabilities. These features \( V \) are then processed through a spatio-temporal encoder \( f_{\text{ST}} \) to capture dynamic interactions, resulting in a comprehensive representation \( E \):

\begin{equation}
V = f_{\text{enc}}(S), \quad E = f_{\text{ST}}(V)
\end{equation}

Subsequently, a two-stage projector is employed to achieve cross-modal alignment. The first stage aligns the spatio-temporally encoded features \( E \) with a textual representation, and the second stage adapts this representation specifically to the event modality, resulting in an event-aligned feature \( E \). Given a user query \( Q \), the large language model \( f_{\text{llm}} \) performs reasoning over \( E \) to generate a response \( A \):

\begin{equation}
E = f_{\text{proj}}(f_{\text{align}}(E)), \quad A = f_{\text{llm}}(E \mid Q)
\end{equation}

EventGPT creates a strong representation of event streams and incorporates task-specific heads to support downstream tasks like object detection and instance segmentation. This flexibility enhances its applicability to a wide range of complex vision tasks.

%EventGPT establishes a robust representation of event streams and integrates task-specific heads to support downstream tasks, such as object detection and instance segmentation. This flexibility broadens its applicability across complex vision tasks.

%Segmenting the event stream into temporal bins structures it 

\iffalse
as a four-dimensional tensor as:

\begin{equation}
\mathcal{E} = \{E_t\}_{t=1}^T, \quad E_t \in \mathbb{R}^{H \times W \times C} 
\end{equation}
where \( T \) is the number of temporal bins, \( H \) and \( W \) denote the spatial dimensions, and \( C \) is the number of channels. 
\fi

\subsection{Spatio-Temporal Information Representation}
\label{sec:Spatiotemporal}
The event stream provides rich spatiotemporal cues, which are crucial for precise temporal and spatial modeling, especially in high-speed motion scenarios to enhance a model’s contextual understanding. The continuous event stream is usually divided into discrete event bins. Each temporal bin \( E_t \) is then transformed by an event encoder \( \mathcal{F}_{\text{enc}} \) as follows:

\begin{equation}
\mathcal{Z} = \{f_{\text{enc}}(E_t)\}_{t=1}^T, \quad \mathcal{Z} \in \mathbb{R}^{T \times S \times D},
\end{equation}
where \( S \) denotes the spatial resolution of the encoded features, and \( D \) represents the feature dimensionality after encoding. To capture salient information across both the temporal and spatial dimensions, we apply independent averaging pooling operations on each dimension.

% Define the max pooling along the temporal dimension \( T \) as \( \mathcal{Z}_T^{\max} \), and similarly, along the spatial dimension \( S \) as \( \mathcal{Z}_S^{\max} \):

% \begin{equation}
% \mathcal{Z}_T^{\max} = \max_{t \in [1, T]} \mathcal{Z}_t, \quad \mathcal{Z}_T^{\max} \in \mathbb{R}^{S \times D}
% \end{equation}

% \begin{equation}
% \mathcal{Z}_S^{\max} = \max_{s \in [1, S]} \mathcal{Z}_s, \quad \mathcal{Z}_S^{\max} \in \mathbb{R}^{T \times D}
% \end{equation}
We define the average pooling operation along the temporal dimension \( T \) as \( \mathcal{Z}_T^{\text{avg}} \):

\begin{equation}
\mathcal{Z}_T^{\text{avg}} = \frac{1}{T} \sum_{t=1}^{T} \mathcal{Z}_t, \quad \mathcal{Z}_T^{\text{avg}} \in \mathbb{R}^{S \times D}.
\end{equation}
Similarly, the average pooling operation \( \mathcal{Z}_S^{\text{avg}} \) along the spatial dimension \( S \) is defined as:

\begin{equation}
\mathcal{Z}_S^{\text{avg}} = \frac{1}{S} \sum_{s=1}^{S} \mathcal{Z}_s, \quad \mathcal{Z}_S^{\text{avg}} \in \mathbb{R}^{T \times D}.
\end{equation}
Finally, we concatenate these max-pooled representations along the feature dimension to form a fused spatiotemporal representation as follows:
\begin{equation}
\overline{\mathcal{Z}} = \text{Concat}(\mathcal{Z}_T^{\max}, \mathcal{Z}_S^{\max}), \quad \overline{\mathcal{Z}} \in \mathbb{R}^{(T + S) \times D}.
\end{equation}
The fused representation \( \overline{\mathcal{Z}} \) is an end-to-end learned spatiotemporal feature from asynchronous events, forming a robust event embedding from various complex scenarios. % When integrated with a large language model (LLM), this spatiotemporal encoding enables contextually-informed response generation by incorporating nuanced temporal and spatial dynamics of event-based scenes.

\subsection{Training Pipelines}
\label{sec:training}
\begin{figure}[t]
  \centering
   \includegraphics[width=\linewidth]{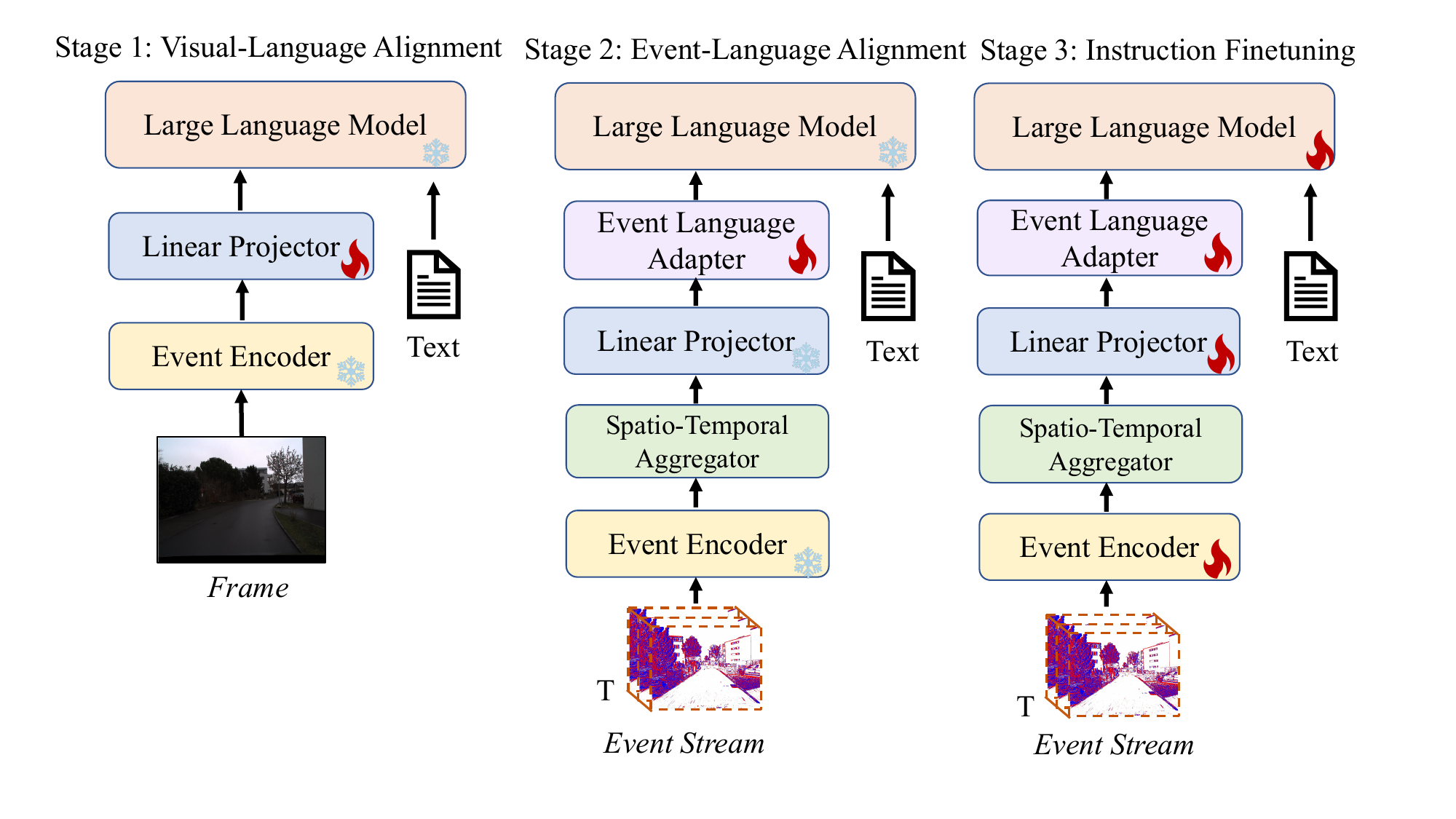}
%    \caption{\textcolor{red}{Our three-stage} pipeline includes: Stage 1 training a Linear Projector for visual-language alignment, Stage 2 adding an Spatio-Temporal
% Aggregator and an Event-Language Adapter for event-language integration, and Stage 3 full fine-tuning of all modules for cohesive multimodal learning.}
   \caption{Our three-stage pipeline consists of (1) training a linear projector to achieve visual-language alignment, (2) incorporating a spatio-temporal aggregator and an event-language adapter to enable event-language alignment, and (3) performing full fine-tuning of all modules to facilitate cohesive multimodal learning.}
   \label{fig:train_pipelines}
   \vspace{-0.30cm}
\end{figure}

\begin{table*}[ht]
\centering
% \caption{Performance comparison on N-ImageNet-Chat dataset and Event-Chat dataset.}
\caption{Performance comparison between EventGPT and five state-of-the-art methods on the N-ImageNet-Chat and Event-Chat datasets across three evaluation metrics. EventGPT consistently demonstrates competent performance on all datasets.}
\label{tab:results} 
\setlength{\tabcolsep}{10pt} 
%\footnotesize
\resizebox{\textwidth}{!}{ 
\begin{tabular}{@{}ccccccccccc@{}}
\toprule
\multirow{2}{*}{\textbf{Models}} & \multirow{2}{*}{\textbf{LLM Backbone}} & \multirow{2}{*}{\textbf{Params}} & \multirow{2}{*}{\textbf{Encoder}} & \multicolumn{3}{c}{\textbf{N-ImageNet-Chat}} & \multicolumn{3}{c}{\textbf{Event-Chat}} \\
\cmidrule(lr){5-7} \cmidrule(lr){8-10}
 & & & & DC & CR & VQA & DC & CR & VQA \\
\midrule
LLaVA-7B-v1.5~\cite{liu2024visual} & Vicuna-v1.5 & 7B & CLIP ViT-L/336PX & 1.54 & 1.07 & 1.88 & 2.20 & 4.04 & 3.26 \\
Qwen2-VL-7B~\cite{wang2024qwen2} & Qwen2 & 7B & OpenCLIP ViT-bigG & 1.74 & 1.46 & 1.91 & 2.38 & 4.02 & 2.91 \\
Intern2VL-8B~\cite{chen2024internvl} & InternLM2.5 & 8B & InternViT-300M & 1.51 & 1.87 & 2.08 & 2.37 & 4.00 & 3.71 \\
Deepseek-vl-7b~\cite{lu2024deepseek} & Deepseek-LLM & 7B & SAM-B + SigLIP-L & 1.52 & 1.45 & 1.84 & 2.41 & 4.10 & 3.37 \\
\rowcolor{verylightgray} \textbf{EventGPT-7B} & Vicuna-v1.5 & 7B & CLIP ViT-L/336PX &2.39 & 2.57 & 2.23 & \textbf{3.52} & 4.09 & \textbf{4.29} \\
\midrule
LLaVA-13B-v1.5~\cite{liu2024visual} & Vicuna-v1.5 & 13B & CLIP ViT-L/336PX & 1.75 & 1.21 & 1.84 & 2.34 & 4.08 & 3.37 \\
\rowcolor{verylightgray} \textbf{EventGPT-13B} & Vicuna-v1.5 & 13B & CLIP ViT-L/336PX & \textbf{2.41} & \textbf{2.81} & \textbf{2.40} & 3.40 & \textbf{4.13} & 4.26 \\
\bottomrule
\end{tabular}
}
\end{table*}

\begin{figure*}[t]
  \centering
   \includegraphics[width=\linewidth]{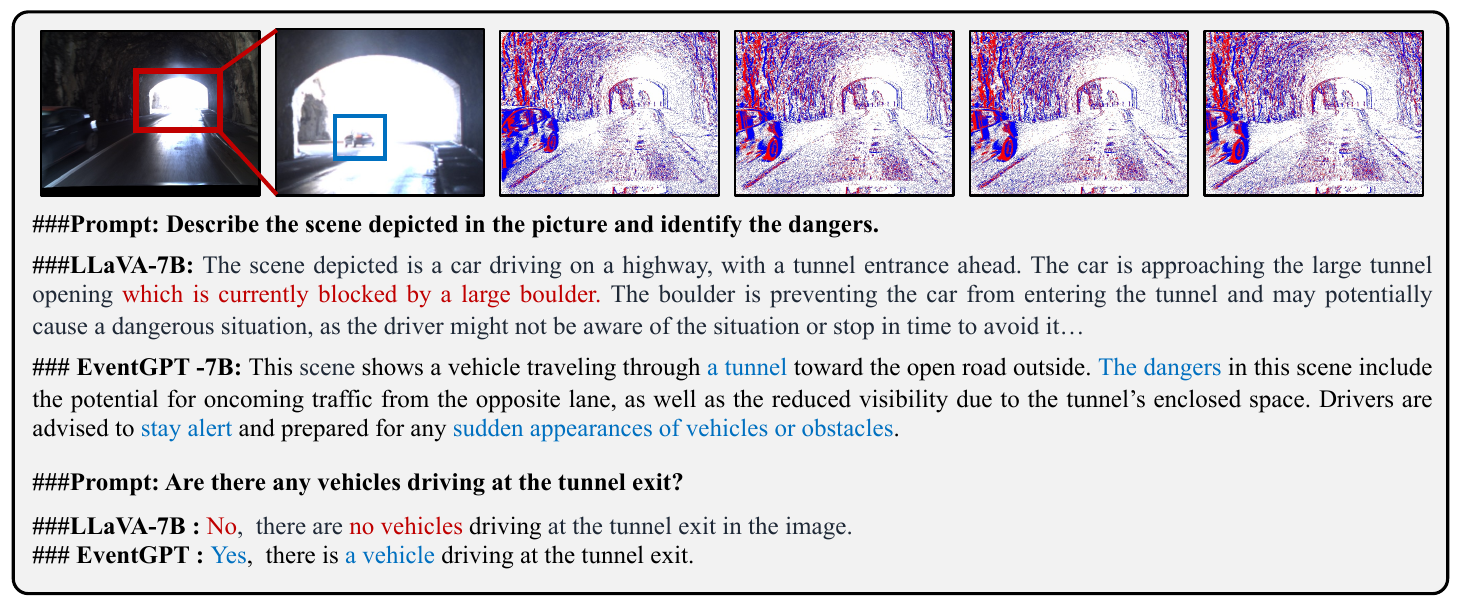}
   \caption{Representative visualization comparison between EventGPT and other open-source MLLMs models under extreme lighting variations. EventGPT demonstrates superior accuracy in recognizing critical scene details, such as vehicle presence at tunnel exits, showcasing its robustness in challenging lighting conditions.}
   \label{fig:tunnel}
   \vspace{-0.30cm}
\end{figure*}

% Extending LLaVA’s two-stage training, EventGPT incorporates an additional alignment stage to adapt visual knowledge specifically for event-domain alignment.
Given the huge domain gap between event and language, we include a visual-language pre-training to initialize our framework. As shown in Fig.~\ref{fig:train_pipelines}, EventGPT adopts a three-phase training framework encompassing visual-language, event-language, and instruction tuning. This structured progression facilitates effective cross-domain alignment, enhancing EventGPT’s capacity for multimodal reasoning and understanding.
%each tailored to distinct objectives that progressively refine its event-domain perception and comprehension. This structured progression facilitates effective cross-domain alignment, enhancing EventGPT’s capacity for event-centric reasoning and multimodal understanding.

\noindent\textbf{Visual-Language Adapter}. In the visual-language alignment stage, we leverage the LLaVA-Pre-train dataset while keeping the event encoder and LLM modules frozen, focusing exclusively on training the linear projector. Visual-language alignment is prioritized due to the relatively small domain differences between image and text, compared to the distinct nature of event and textual data. This setup establishes initial cross-modal alignment between vision and language, enabling the LLM to interpret scene information and providing a foundational multimodal understanding.

\noindent\textbf{Event-Language Adapter}. In the event-language alignment stage, we leverage our constructed N-ImageNet-Chat dataset, with the event encoder, linear projector, and LLM modules kept frozen, focusing exclusively on training the event-language adapter module. This stage aims to bridge the domain gap between event and language modalities. By aligning event embeddings with the LLM’s representational space, the model’s ability to interpret event streams and generate descriptive narratives is significantly enhanced.

\noindent\textbf{Instruction Tuning}. In the instruction tuning stage, all trainable parameters, including the event encoder, linear projector, event-language adapter, and LLM modules, are unfrozen to enable supervised fine-tuning. Trained on our high-quality Event-Chat question-answering dataset, this stage further enhances the model’s understanding and generation capabilities for event stream data. This structured, three-phase training framework progressively narrows the domain gap between event and language modalities.

%Through this progressive three-phase training framework, the model incrementally bridges the domain gap between event and language modalities, ultimately enhancing its capacity for event-centric reasoning and generating responses grounded in an event-informed multimodal understanding.

\section{Experiment}
\subsection{Experiment Settings}
\noindent \textbf{Implementations}. We integrate the N-ImageNet\cite{kim2021n}, Dsec\cite{gehrig2021dsec}, and E2VID\cite{Rebecq19cvpr} datasets and synthesized multimodal conversational datasets, N-ImageNet-Chat and Event-Chat, using a GPT-assistant approach to support the three-stage model training. Our LLM backbone is based on the vicuna-v1.5 version. The linear projector is a two-layer MLP, and the event-language adapter is a linear layer. All models, including the 7B and 13B parameter versions, are trained on 8 NVIDIA RTX 6000 Ada GPUs. During the visual-language alignment stage, we use 558k data, with a learning rate of $2 \times 10^{-4}$ and a batch size of 64. For the event-language alignment stage, we use 1000k data, maintaining the same learning rate and batch size. In the instruction-supervised fine-tuning stage, we use 120k data, setting the learning rate to $2 \times 10^{-5}$ with a batch size of 32. 

\noindent\textbf{Evaluation Metrics.} Following the evaluation strategies of LLaVA~\cite{liu2024visual} and VideoGPT~\cite{maaz2023video}, we rigorously assess EventGPT’s performance against state-of-the-art MLLMs using three core metrics within a zero-shot question-answering setting. Our benchmark, specifically designed for text generation from event streams, is based on the test sets of N-ImageNet-Chat and Event-Chat datasets, which contain question-answer pairs to evaluate the model’s abilities in detailed description, complex reasoning, and visual question answering. 
% The generated responses are quantitatively scored on a 1-5 scale across three dimensions: (i) Detailed Captioning (DC), assessing the accuracy and completeness of descriptive outputs to capture critical event details; (ii) Complex Reasoning (CR), evaluating the model’s capacity for multi-step reasoning with logically inferred responses; and (iii) Visual Question Answering (VQA), measuring precision and contextual comprehension in answering specific visual queries.
Generated responses are quantitatively scored on a 1-5 scale across three dimensions: (i) Detailed Captioning (DC), assessing the accuracy and completeness of descriptions to capture key event details; (ii) Complex Reasoning (CR), evaluating the model’s capacity for multi-step reasoning and logical inference; and (iii) Visual Question Answering (VQA), measuring precision and contextual understanding in responding to specific queries.

% \subsection{Main Results}
\subsection{Comparison with State-of-the-Art Models}

\begin{figure*}[t]
  \centering
   \includegraphics[width=\linewidth]{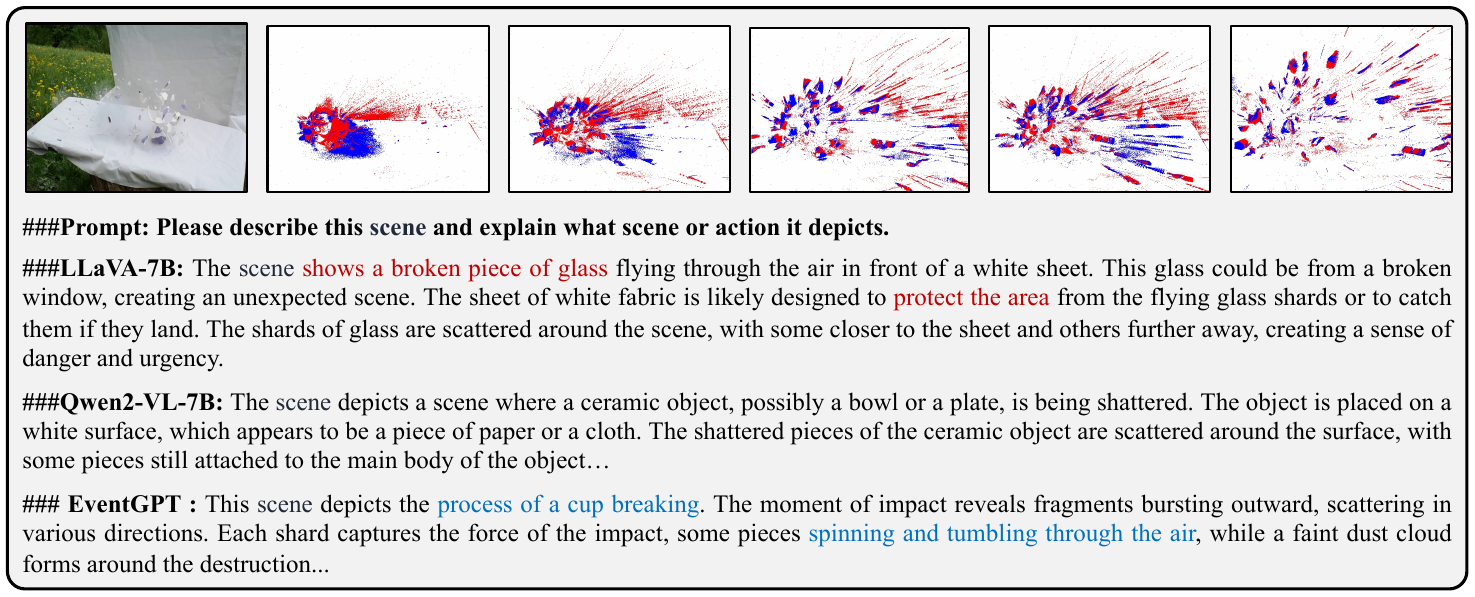}
   \caption{Representative comparison of EventGPT and other open-source multimodal language models (MLLMs) in high-speed motion scenarios, demonstrating EventGPT’s enhanced capability to capture spatiotemporal details and interpret motion characteristics, including fragment trajectories and impact effects.}
   \label{fig:cup}
   \vspace{-0.30cm}
\end{figure*}

\bfsection{Quantitative Results}
To verify the effectiveness of our EvenGPT, we present the quantitative evaluation results of our proposed EventGPT model (available in 7B and 13B versions) in Table~\ref{tab:results}. To maintain an end-to-end architecture and ensure fair comparison, we represent the event stream as an event image under identical input conditions. We compare our model with state-of-the-art MLLMs, including Qwen2-VL~\cite{wang2024qwen2}, DeepSeek-VL~\cite{lu2024deepseek}, Intern2VL~\cite{chen2024internvl}, and LLaVA~\cite{liu2024visual}. The results demonstrate that EventGPT exhibits strong capabilities in event understanding, which is mainly attributed to our three-stage training framework that gradually bridges the significant gap between events and language. 

\bfsection{Qualitative Analysis}
In addition, we conduct qualitative assessments on selected challenging scenarios from the Event-Chat test set, specifically those involving extreme lighting conditions or high-speed object dynamics (see Fig.~\ref{fig:tunnel} and Fig.~\ref{fig:cup}). The results illustrate that our proposed EventGPT consistently outperforms other models in these difficult settings, demonstrating robust event-driven scene understanding capabilities. For high dynamic range scenarios, such as the tunnel scene in Fig.~\ref{fig:tunnel}, EventGPT successfully identifies the tunnel environment and issues relevant hazard warnings. Other models, however, frequently misinterpret the scene or fail to detect the tunnel due to the extreme light changes at the exit point.
% This scenario highlights the unique advantages of event-based data in handling drastic illumination shifts, as EventGPT effectively captures the high-exposure elements at the tunnel entrance and generates timely alerts. 
% This scenario exemplifies a typical application for event-based data, which excels in handling drastic illumination shifts. 
In such conditions, EventGPT demonstrates a unique advantage by effectively capturing high-exposure elements at the tunnel exit and generating timely alerts.
For high-speed motion scenarios, such as a cup breaking in Fig.~\ref{fig:cup}, EventGPT demonstrates advanced spatio-temporal understanding by effectively tracking and capturing the dynamic rupture sequence of the object. In contrast, LLaVA-7B misinterprets the scene with an inaccurate description, while Qwen2-VL-7B lacks spatio-temporal comprehension, producing a static and incomplete depiction.

% EventGPT demonstrates advanced spatio-temporal comprehension by accurately tracking and capturing the object’s dynamic rupture sequence. Other models either fail to recognize the specific scene or are unable to capture the rapid event accurately.

% These qualitative results emphasize EventGPT’s superior performance in understanding complex scenarios, particularly when compared to traditional image-based multimodal large language models.

\subsection{Ablation Study}
\noindent\textbf{Contribution of Each Component}. 
% To explore the impact of each component on the final performance, we adopt a training method similar to the LLaVA series models as our baseline, with the key difference that we use event images for the second-stage instruction fine-tuning on the Event-Chat dataset. 
To explore the impact of each component, we first construct a baseline similar to LLaVA except adopting event frames in the second training stage.
% adopt a training method similar to the LLaVA series models as our baseline, with the key difference that we use event images for the second-stage instruction fine-tuning on the Event-Chat dataset.
Two variants are constructed by incorporating our event-language adapter or spatio-temporal aggregator. 
As shown in Table~\ref{tab:ablation}, our proposed EventGPT achieves higher performance compared to the baseline and variants. Compared to the baseline, incorporating the spatio-temporal aggregator results in respective improvement of 2.35\%, 1.26\%, and 1.20\% on the DC, CR, and VQA metrics, significantly enhancing the model's capability of capturing temporal dynamics and spatial relationships within event streams, which facilitates to better interpret complex motion patterns and scene changes. 
% This leads to more accurate responses in tasks requiring temporal reasoning and context awareness.
Additionally, the event-language adapter brings improvements of 3.24\%, 2.02\%, and 2.41\% on the DC, CR, and VQA metrics respectively over the baseline. This highlights the superiority of our training pipeline. We extend the traditional two-stage paradigm to a three-stage fine-grained progress. By progressively bridging the large cross-modal gap, our model achieves more accurate event stream perception and contextual understanding.
\begin{table}[t]
\centering
\footnotesize
% \caption{Comparison of EventGPT-7B variants without the event-language adapter or spatio-temporal aggregator.}
\caption{Performance impact of each component in EventGPT through ablation study. The results demonstrate that both the spatio-temporal aggregator and the event-language adapter contribute significant performance improvements over the baseline.}
\label{tab:ablation}
\resizebox{\columnwidth}{!}{ 
\begin{tabular}{c@{\hspace{1pt}}c@{\hspace{1pt}}cccc}
\toprule
\multirow{2}{*}{\textbf{Mode Type}} & \multirow{2}{*}{\makecell{\textbf{Event} \\ \textbf{Language}}} & \multirow{2}{*}{\makecell{\textbf{Spatio} \\ \textbf{Temporal}}} & \multicolumn{3}{c}{\textbf{Event-Chat}} \\
\cmidrule(lr){4-6}
& & & \textbf{DC} & \textbf{CR} & \textbf{VQA} \\
\midrule
Baseline & \xmark & \xmark & \makecell[lt]{3.40} & \makecell[lt]{3.97} & \makecell[lt]{4.15} \\
Variant A & \cmark & \xmark & \makecell[lt]{3.48\tiny{(+2.35\%)}} & \makecell{4.02\tiny{(+1.26\%)}} & \makecell{4.20\tiny{(+1.20\%)}} \\
Variant B & \xmark & \cmark & \makecell{3.51\tiny{(+3.24\%)}} & \makecell{4.05\tiny{(+2.02\%)}} & \makecell{4.25\tiny{(+2.41\%)}} \\
\rowcolor{verylightgray} \textbf{Ours} & \cmark & \cmark & \makecell{\textbf{3.52}\tiny{(+3.53\%)}} & \makecell{\textbf{4.09}\tiny{(+3.02\%)}} & \makecell{\textbf{4.29}\tiny{(+3.37\%)}} \\
\bottomrule
\end{tabular}
}
\end{table}

\noindent\textbf{Influence of Temporal Aggregation Size.} To explore the effect of temporal aggregation size, we conduct an analysis to determine the optimal temporal window number \( N_w \) within the spatio-temporal aggregator. \( N_w \) represents the number of temporal bins used to divide a fixed observation period, with each bin aggregating events within its respective interval. Fewer windows may fail to capture fine-grained temporal details, while more windows can capture richer temporal cues but lead to sparser event distribution per bin, making it challenging to model context information effectively. In other words, a small number of windows fail to capture sufficient scene dynamics, while an excessively large number of windows lead to the loss of scene context.
% Thus, identifying an optimal temporal window number is essential for effective spatiotemporal modeling.
The quantitative results in Table~\ref{tab:time_windows} reveal that our framework is robust to a wide range of temporal aggregation sizes. 
Furthermore, a temporal window number of 5 yields the best overall performance. Accordingly, we set \( N_w =5\) in the spatio-temporal aggregator by default.
% To explore the effect of temporal aggregation size, we conduct experiments to determine the optimal number of the temporal window within its spatio-temporal aggregator (see Table~\ref{tab:time_windows}). We configure the aggregator with various temporal window numbers (3, 5, 7, 9, 11, and 13) to examine their effect on spatio-temporal modeling. In this study, the temporal window length refers to the duration of each interval over which events are aggregated within the total observation period. Larger temporal windows capture richer temporal cues from the event stream but significantly reduce the number of events per window, which can limit the encoder’s ability to extract meaningful features. Thus, finding the optimal temporal window length is essential for effective spatiotemporal modeling. From Table~\ref{tab:time_windows}, our quantitative results demonstrate that EventGPT’s performance remains relatively stable across different temporal window lengths. However, a window length of 5 yielded the best performance overall. Consequently, we set the temporal window length of EventGPT’s spatio-temporal aggregator to 5 to optimize performance.

\begin{table}[t]
\centering
\footnotesize
\caption{Performance comparison of EventGPT-7B across different temporal window numbers \( N_w \). The results demonstrate that our framework is robust to a wide range of window numbers. However, both too few and too many windows adversely affect performance, with the optimal result achieved at \( N_w = 5 \).}
\label{tab:time_windows}
%\resizebox{\columnwidth}{!}{ 
\setlength{\tabcolsep}{2.80mm}{
\begin{tabular}{ccccccc}
\toprule
\multirow{2}{*}{\centering \textbf{\( N_w \)}} & \multicolumn{6}{c}{\textbf{Event-Chat}} \\
\cmidrule(lr){2-7}
 & \textbf{3} & \textbf{5} & \textbf{7} & \textbf{9} & \textbf{11} & \textbf{13}\\
\midrule
DC & 3.47 & 3.52 & \textbf{3.56} & 3.41 &3.36 &3.32\\
CR & 4.03 & \textbf{4.09} & 4.05 & 3.96 &3.97 &3.91\\
VQA & 4.21 & \textbf{4.29} & 4.17 & 4.12 &4.08 &4.03\\
\bottomrule
\end{tabular}}
\end{table}

\subsection{Downstream Applications}
In this section, we assess the effectiveness of EventGPT, our proposed multimodal large language model (MLLM) tailored for event stream understanding, on two downstream tasks, object detection and instance segmentation. To support these tasks, we integrate GroundingDINO \cite{liu2023grounding} and GroundedSAM \cite{ren2024grounded} as task-specific heads, facilitating the adaptation of EventGPT to object detection and instance segmentation, respectively.

\begin{figure}[t]
  \centering
   \includegraphics[width=\linewidth]{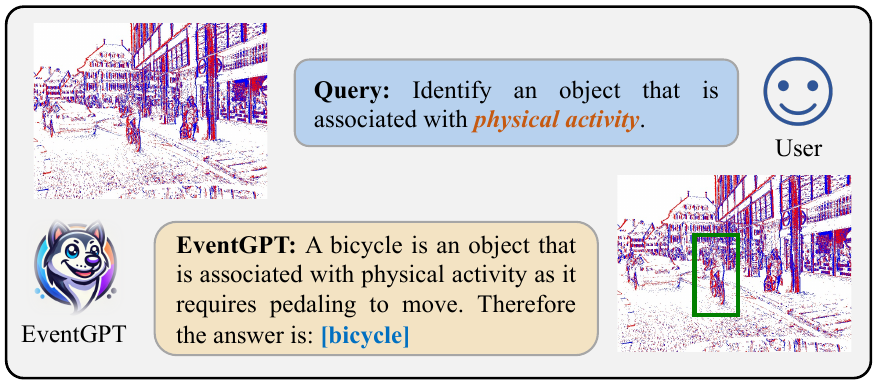}
   \caption{Object detection~\cite{liu2023grounding} using events and text from our EventGPT. Note that, our EventGPT enables the generation of high-quality text for the open-set object detection task.}
   \label{fig:detection}
\end{figure}

\noindent\textbf{Object Detection}. To evaluate the generalization ability of EventGPT on object-level understanding and reasoning, we integrate it with GroundingDINO~\cite{liu2023grounding}, leveraging its open-set object detection capabilities. 
For each question, EventGPT generates textual descriptions of potential objects in the event streams, which are subsequently used by GroundingDINO to localize the corresponding objects within the data. 
As illustrated in Fig.~\ref{fig:detection}, EventGPT demonstrates strong object-level reasoning ability, accurately inferring and localizing target objects in the scenario. This synergy between EventGPT and GroundingDINO showcases its inferential object detection ability and potential applications in domains like autonomous driving and embodied intelligence.
% This synergy between EventGPT and GroundingDINO not only showcases the model's ability of performing inferential object detection but also highlights its potential applicability in domains requiring precise object understanding, such as autonomous driving and embodied intelligence.
% To evaluate the robustness of EventGPT in object-level understanding and reasoning, we bridge it with GroundingDINO \cite{liu2023grounding} to test its effectiveness in object detection scenarios. As shown in Fig. \ref{fig:detection}, EventGPT demonstrates object-level reasoning capabilities, accurately localizing the target objects through inference and achieving robust open-set object detection. The capacity for inferential object detection broadens the potential applications of EventGPT, making it suitable for scenarios requiring robust object understanding, such as autonomous driving and embodied intelligence.

\begin{figure}[t]
  \centering
   \includegraphics[width=\linewidth]{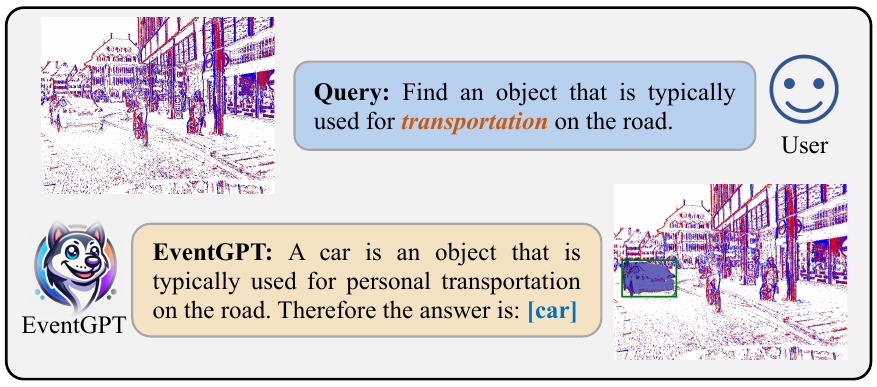}
   \caption{Instance segmentation \cite{ren2024grounded} leveraging events and text generated by our EventGPT. Our EventGPT is capable of producing high-quality text descriptions for open-set instance segmentation tasks.}
   \label{fig:segmentation}
   \vspace{-5mm}
\end{figure}

\noindent\textbf{Instance Segmentation}. To validate EventGPT’s scalability on complex dense prediction tasks, we integrate it with GroundedSAM~\cite{ren2024grounded} for instance segmentation on event streams. EventGPT provides semantic cues to guide GroundedSAM in generating precise instance masks, segmenting individual objects in the scene. The qualitative results in Fig.~\ref{fig:segmentation} demonstrate our approach’s efficacy in achieving accurate segmentation under the challenging conditions inherent to event-based data. This affirms EventGPT’s adaptability and potential in fine-grained scene understanding tasks.
% These affirm EventGPT’s adaptability and potential to enhance the performance of tasks requiring fine-grained scene understanding.
% Through the integration of EventGPT with the GroundedSAM head, precise instance segmentation is achieved, allowing for the generation of detailed and accurate instance masks on event-based data. The segmentation results are shown in Fig. \ref{fig:segmentation}, demonstrating the model’s effectiveness. These visualizations demonstrate EventGPT’s adaptability in applying event-based understanding to distinct vision tasks, showing its functionality across object detection and instance segmentation tasks.

% \begin{table}[ht]
% \centering
% \footnotesize
% \caption{Performance Comparison of EventGPT-7B on Different Mode Types and Time Steps}
% \label{tab:time_windows}
% \resizebox{\columnwidth}{!}{
% \begin{tabular}{cccccc}
% \toprule
% \multirow{2}{*}{\centering \textbf{Mode Type}}  & \multirow{2}{*}{\centering \textbf{Time Steps}} & \multicolumn{4}{c}{\textbf{Event-Chat}} \\
% \cmidrule(lr){3-6}
% & & \textbf{3} & \textbf{5} & \textbf{7} & \textbf{9} \\
% \midrule
% \multirow{3}{*}{EventGPT-7B} & DC & 3.47 & 3.52 & 3.56 & 3.41 \\
%                              & CR & 4.03 & \textbf{4.09} & 4.05 & 3.96 \\
%                              & VQA & 4.21 & \textbf{4.29} & 4.17 & 4.12 \\
% \bottomrule
% \end{tabular}
% }
% \end{table}

\section{Conclusion}
% This paper introduces EventGPT, the first multimodal large language model tailored for event stream understanding. The model consists of an event encoder, a spatio-temporal aggregator, a linear projector, an event-language adapter, and an LLM. It utilizes a three-stage training pipeline to progressively bridge the huge domain gap between event streams and language for scene perception and description. The comprehensive evaluation demonstrates the superiority of our \name{} on event scene understanding.
% Additionally, we provide two large-scale event-text datasets, N-ImageNet-Chat and Event-Chat, to advance the cross-modal alignment research in the community. 
This paper introduces EventGPT, the first multimodal large language model tailored for event stream understanding. The model comprises an event encoder, a spatio-temporal aggregator, a linear projector, an event-language adapter, and an LLM. It employs a three-stage training pipeline to progressively bridge the significant domain gap between event streams and language for scene perception and description. Comprehensive evaluations demonstrate the superiority of our method in event scene understanding. Additionally, we build two large-scale event-text datasets, N-ImageNet-Chat and Event-Chat, to advance cross-modal alignment research in the community.

% We hope this work inspires advancements in LLM architectures for event stream processing, the creation of robust Event-Text datasets, and further research to enhance multimodal models’ perceptual capabilities.

% \noindent \textbf{Limitation.} Our EventGPT currently processes only event data, making it challenging to achieve high-accuracy inference in static or slow-motion scenes. In the future, we plan to expand EventGPT to jointly process frames and events.
\newpage
{
    \small
    \bibliographystyle{ieeenat_fullname}
    \bibliography{main}
}

% \input{author-kit-CVPR2025-v3-latex-/sec/X_suppl}

\end{document}